\documentclass[conference]{IEEEtran}

\usepackage{cite}
\usepackage{amsmath,amssymb,amsfonts}
\usepackage{algorithmic}
\usepackage{graphicx}
\usepackage{textcomp}
\usepackage[table]{xcolor}
\usepackage{booktabs}
\usepackage{bm}
\usepackage{multirow}
\usepackage{makecell}
\usepackage{comment}
\usepackage{tabularx}
\usepackage[T1]{fontenc}
\newcolumntype{C}{>{\centering\arraybackslash}X}
\usepackage{enumitem}

\usepackage{pifont}

\newcommand{\cmark}{\ding{51}} 
\newcommand{\xmark}{\ding{55}} 

\usepackage{pdfpages}

\definecolor{yeoVis}{HTML}{781286}        
\definecolor{yeoSomMot}{HTML}{4682B4}     
\definecolor{yeoDorsAttn}{HTML}{00760E}   
\definecolor{yeoSalVentAttn}{HTML}{C43AFA}
\definecolor{yeoLimbic}{HTML}{DCCA4F}     
\definecolor{yeoCont}{HTML}{E69422}       
\definecolor{yeoDefault}{HTML}{CD3E4E}    

\newcommand{\meanstd}[2]{%
  \ensuremath{#1{\scriptstyle\mkern1mu\pm\mkern1mu#2}}%
}

\newcommand{\bestmeanstd}[2]{%
  \ensuremath{\mathbf{#1}%
  {\scriptstyle\mkern1mu\boldsymbol{\pm}\mkern1mu\mathbf{#2}}}%
}
\definecolor{refpink}{HTML}{E66482}

\usepackage[
    colorlinks=true,
    linkcolor=black,  
    citecolor=refpink,  
    urlcolor=refpink
]{hyperref}

\begin{document}

\title{MovieSTAGE: Scene, Transition, and Global Encoding for Movie-fMRI ADHD Classification}
\author{
Boseong Kim\textsuperscript{1},
Haejun Chung\textsuperscript{1},
and Ikbeom Jang\textsuperscript{2}\\[0.5em]
\textsuperscript{1}Hanyang University, Seoul, Republic of Korea\\
\textsuperscript{2}Hankuk University of Foreign Studies, Yongin, Republic of Korea\\
Correspondence: ijang@hufs.ac.kr; haejun@hanyang.ac.kr
}

\maketitle

\begin{abstract}

Naturalistic movie-fMRI provides a shared, temporally structured probe of brain dynamics, yet predictive models commonly rely on whole-run functional connectivity (FC) or temporally generic representations that are not aligned with narrative events. We introduce MovieSTAGE (Scene, Transition, and Global Encoding), a multiscale framework that combines hypergraph-structured FC-profile organization within scenes, unsigned FC-profile differences across adjacent scenes, and whole-movie FC. We evaluated 260 participants from the CMI-HBN \emph{Despicable Me} cohort on case--control, ADHD-subtype, and three-class classification using 10 repetitions of stratified five-fold cross-validation, complete out-of-fold (OOF) predictions, and paired subject-cluster bootstrap and permutation tests. MovieSTAGE achieved AUROCs of 0.69, 0.73, and 0.75 and balanced accuracies of 67.6\%, 69.8\%, and 58.3\%, respectively, yielding the highest mean point estimates among the evaluated methods. On the three-class task, the full model outperformed all two-branch variants, the HGNN scene encoder outperformed MLP, GAT, and BNT alternatives under matched settings, and the human-annotated partition outperformed duration-matched random and fixed-count GSBS controls. These controlled results support incremental predictive value from event-aligned scene and transition representations when combined with whole-movie FC in this cohort. Post-hoc model-derived analyses generated network-level hypotheses involving frontoparietal and default-mode systems.

\end{abstract}
\begin{IEEEkeywords}
movie-fMRI, hypergraph neural network, attention-deficit/hyperactivity disorder, functional connectivity
\end{IEEEkeywords}


\vspace{-1mm}
\section{Introduction}
\vspace{-1mm}

Attention-deficit/hyperactivity disorder (ADHD) is clinically and neurobiologically heterogeneous, and neuroimaging studies have implicated altered interactions among the frontoparietal control network (FPN), default-mode network (DMN), and other large-scale systems~\cite{tegelbeckers2015altered,fair2010atypical,
mills2018adhd,duffy2021increased,shappell2021children,yao2014functional}. Predictive functional magnetic resonance imaging (fMRI) studies commonly summarize these alterations using resting-state or whole-run functional connectivity (FC), which provides a subject-level representation but collapses within-run temporal organization~\cite{shen2017using}. Naturalistic movie-fMRI presents participants with a shared, temporally structured stimulus engaging perceptual, social, affective, and narrative processes, enabling ADHD-related differences to be examined during continuous experience rather than isolated trials~\cite{salmi2020adhd}.

Recent fMRI--AI approaches extend connectome-based prediction beyond vectorized FC to graph, hypergraph, and spatiotemporal representations~\cite{cao2021temporal}. Hypergraph formulations encode higher-order relations among brain regions~\cite{xiao2019multi, feng2019hypergraph,wang2022equivariant}, whereas dynamic graph and attention-based models learn from sliding-window FC, region-of-interest (ROI) time series, or whole-connectome topology~\cite{ICLR2025_81f19c0e, thapaliya2025dsam,wang2025stnagnn,ma2026brainhgt}. However, they operate on whole runs, fixed windows, or latent snapshots rather than narrative events, leaving it unclear whether predictive gains reflect temporal variation or event alignment.

During continuous narratives, cortical activity patterns remain relatively stable within events and shift at event boundaries~\cite{baldassano2017discovering,lee2024relationship}. Dynamic-FC studies likewise indicate that time-varying network configurations carry information beyond a run-averaged connectome~\cite{savva2019assessment,preti2017dynamic}. Event-aligned modeling nevertheless faces two challenges: limited temporal support for scene-wise FC and potential loss of global context. We therefore hypothesize that three representations provide complementary information: set-wise within-scene organization, adjacent-scene reconfiguration, and whole-movie FC. ROI FC-profile similarity supports scene-specific hypergraphs without relying on isolated short-window edges; absolute differences between adjacent scene-wise FC profiles quantify reconfiguration without assuming a common signed direction across participants; and whole-movie FC preserves global context.


We introduce \textbf{MovieSTAGE} (\textbf{S}cene, \textbf{T}ransition, \textbf{a}nd \textbf{G}lobal \textbf{E}ncoding) for movie-fMRI ADHD prediction. For each scene, MovieSTAGE constructs a subject-specific ROI hypergraph from FC-profile similarity and aggregates scene embeddings across the movie. A second branch encodes adjacent-scene FC-profile changes, while a third represents whole-movie FC.

Our contributions are threefold:
\begin{enumerate}[label=(\roman*), leftmargin=*, itemsep=0.2em]

    \item We introduce \textbf{MovieSTAGE}, a scene--transition--global framework  combining scene-wise hypergraphs, adjacent-scene connectivity-profile reconfiguration, and whole-movie FC for movie-fMRI prediction.

    \item We evaluate MovieSTAGE across three CMI-HBN ADHD tasks. MovieSTAGE yields the highest mean AUROC and BACC point estimates among the evaluated baselines, while ablations and matched segmentation controls indicate partially complementary representations and a benefit of narrative alignment.

    \item We provide multiscale model-derived analyses highlighting higher-order scene organization across ADHD-relevant large-scale systems and DMN-centered adjacent-scene reconfiguration patterns.
    
\end{enumerate}

\begin{figure*}[!t]
    \centering
    \includegraphics[width=0.95\textwidth]{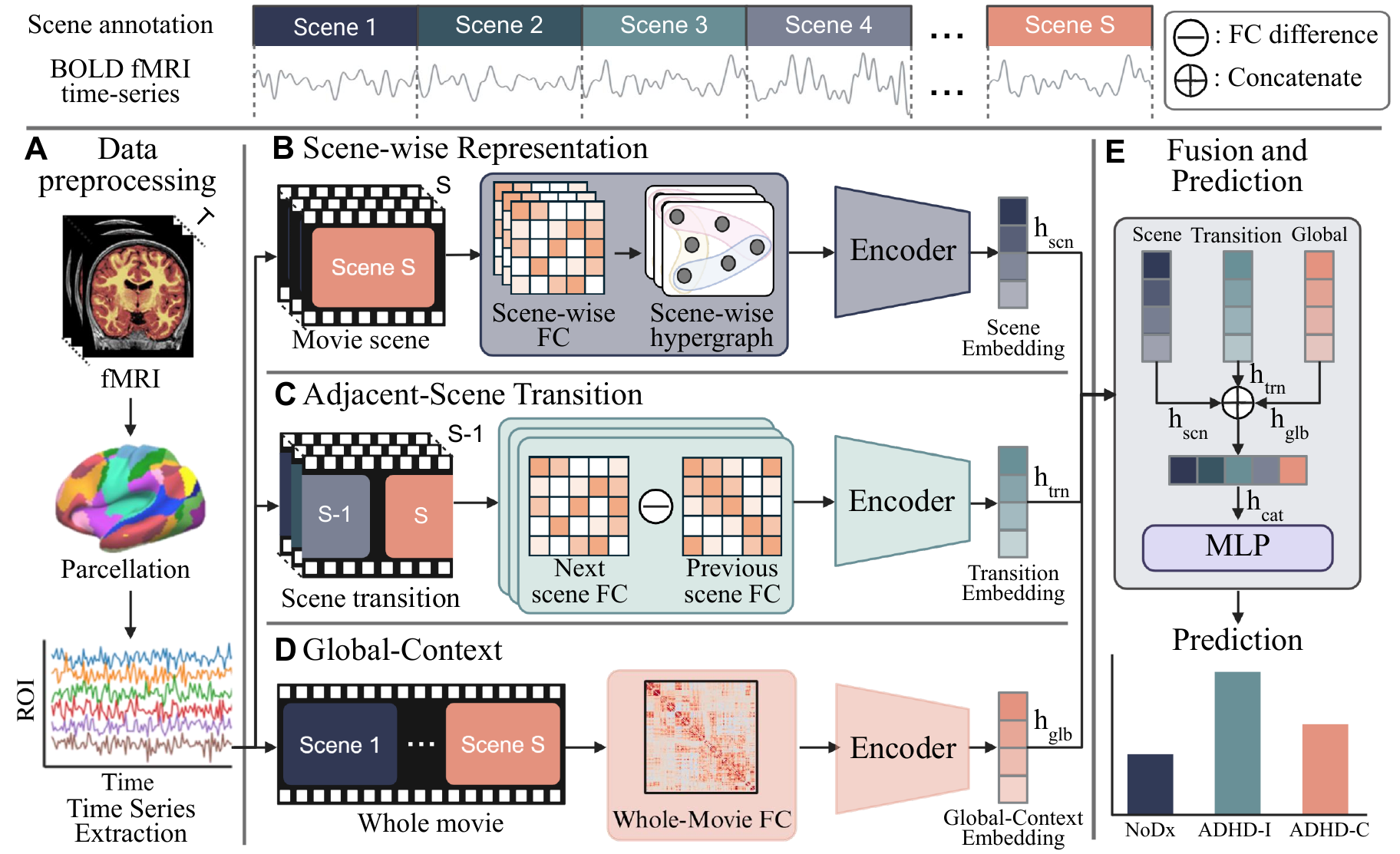}
    \caption{Overview of MovieSTAGE, a scene--transition--global fusion network. Preprocessed movie-fMRI time series are parcellated into SC-100 ROI signals and segmented by movie scene boundaries. The scene branch encodes scene-wise FC-profile hypergraphs, the transition branch encodes adjacent-scene unsigned \(\Delta\mathrm{FC}\), and the global branch summarizes whole-movie FC. The three embeddings are concatenated for subject-level prediction.}
    \label{fig:main}
    \vspace{-4mm}
\end{figure*}

\section{Methods}
\label{sec:methods}

Let \(\mathbf{X}_i\in\mathbb{R}^{R\times T}\) denote the parcellated movie-fMRI time series of subject \(i\), with \(R\) ROIs and \(T\) time points. The movie is divided into \(S\) scenes using predefined event boundaries; \(\mathbf{X}_{i,s}\) denotes scene \(s\), and boundary \(q\) is the transition from scene \(q\) to \(q+1\). MovieSTAGE contains three branches: scene-wise hypergraph (\(\mathrm{scn}\)), adjacent-scene transition (\(\mathrm{trn}\)), and global FC (\(\mathrm{glb}\)). Unless otherwise specified, FC denotes a Fisher-\(z\)-transformed correlation matrix derived from a Ledoit--Wolf shrinkage covariance estimate~\cite{ledoit2004well}. The estimator was applied independently to each subject and temporal block, including each narrative scene and the complete movie. The shrinkage covariance matrix was converted to a correlation matrix by variance normalization, and diagonal entries were set to zero after applying the Fisher transformation to the off-diagonal entries. The MovieSTAGE architecture is shown in \autoref{fig:main}.

\subsection{Scene-wise Hypergraph Representation}
\label{sec:scene_hypergraph_representation}

For each subject and scene, we compute shrinkage-regularized FC and use each ROI row as a scene-specific FC-profile descriptor:
\begin{equation}
\mathbf{FC}^{\mathrm{scn}}_{i,s} = \mathcal{Z}\!\left(\operatorname{Corr}_{\mathrm{LW}}(\mathbf{X}_{i,s}) \right), \quad
\mathbf{x}^{\mathrm{scn}}_{i,s,r} = \mathbf{FC}^{\mathrm{scn}}_{i,s}[r,:].
\label{eq:scene_fc}
\end{equation}

Here, \(\operatorname{Corr}_{\mathrm{LW}}(\cdot)\) denotes the correlation matrix obtained by variance-normalizing a Ledoit--Wolf shrinkage covariance estimate, and \(\mathcal{Z}(\cdot)\) applies the Fisher \(r\)-to-\(z\) transformation to off-diagonal entries. Because scene-wise FC is estimated from limited temporal samples, Ledoit--Wolf shrinkage was used to regularize covariance estimation before constructing similarity-based FC-profile hyperedges.

A scene-specific hypergraph is constructed from the cosine-similarity matrix \(\mathbf{S}_{i,s}\), where
\(\mathbf{S}_{i,s}[r,r']\) is the cosine similarity between \(\mathbf{x}^{\mathrm{scn}}_{i,s,r}\) and
\(\mathbf{x}^{\mathrm{scn}}_{i,s,r'}\). For anchor ROI \(r\), one
hyperedge connects \(r\) with its top-\(K\) most similar ROIs:
\begin{equation}
    e_{i,s,r}=\{r\}\cup \operatorname{TopK}_{r'\neq r}\left(\mathbf{S}_{i,s}[r,r']\right).
\label{eq:scene_hyperedge}
\end{equation}
This yields \(E=R\) hyperedges per scene. The incidence matrix \(\mathbf{H}^{\mathrm{scn}}_{i,s}\in\{0,1\}^{R\times E}\) is defined by \(H^{\mathrm{scn}}_{i,s}[r',r]=1\) when ROI \(r'\) belongs to the hyperedge anchored at ROI \(r\). The hyperedge attribute matrix \(\mathbf{A}^{\mathrm{scn}}_{i,s}\in\mathbb{R}^{E\times1}\) stores, for each anchor, the mean cosine similarity to its \(K\) selected members. Hypergraphs are constructed only from subject-specific scene-wise FC profiles and do not use diagnostic labels. The initial node-feature matrix is \(\mathbf{Z}^{(0)}_{i,s}=\mathbf{FC}^{\mathrm{scn}}_{i,s}\).

The scene branch applies a shared hypergraph encoder with learnable hyperedge weights. Omitting subject and scene indices, the layer update is
\begin{equation}
    \mathbf{Z}^{(l+1)} = \sigma\!\left((\mathbf{D}_{v}^{(l)})^{-\frac{1}{2}}\mathbf{H}\mathbf{W}_{e}^{(l)}\mathbf{D}_{e}^{-1}\mathbf{H}^{\top}(\mathbf{D}_{v}^{(l)})^{-\frac{1}{2}}\mathbf{Z}^{(l)}\mathbf{W}^{(l)}\right),
\label{eq:hgconv}
\end{equation}
where \(\mathbf{Z}^{(l)}\) is the node embedding matrix, \(\mathbf{W}^{(l)}\) is a node-projection matrix, \(\mathbf{W}_{e}^{(l)}=\operatorname{diag}(\mathbf{w}^{(l)})\), \(D_{e,e}=\sum_v H_{v,e}\), \(D^{(l)}_{v,v}=\sum_e H_{v,e}w_e^{(l)}\), and \(\sigma\) denotes the rectified linear unit (ReLU) activation. The hyperedge weight is

\begin{equation}
w_e^{(l)}=\operatorname{Softplus}\!\left(f_{\mathrm{edge}}\left([\bar{\mathbf{z}}^{(l)}_e,a_e]\right)\right)+\epsilon,
\label{eq:learnable_edge_weight}
\end{equation}
with \(\bar{\mathbf{z}}^{(l)}_e\) denoting the mean embedding of nodes incident to hyperedge \(e\), and \(a_e\) the corresponding scalar hyperedge attribute. Here, \(f_{\mathrm{edge}}\) is a multilayer perceptron (MLP) applied to each hyperedge, and \(\epsilon>0\) sets a positive lower bound on the learned weight. This incidence-matrix aggregation captures multi-ROI scene-level organization \cite{xiao2019multi,feng2019hypergraph}. Final ROI embeddings are summarized by mean, maximum, and standard-deviation pooling, projected to scene embeddings, and aggregated across scenes with attention pooling to produce \(\mathbf{h}^{\mathrm{scn}}_i\).

\subsection{Adjacent-Scene Transition Representation}
\label{sec:boundary_transition_features}

Event boundaries in naturalistic fMRI are associated with transitions between relatively stable activity patterns and cortical pattern shifts \cite{baldassano2017discovering,lee2024relationship}. Consistent with dynamic-FC analyses of time-varying network configuration, we encode adjacent-scene reconfiguration magnitude without assuming a consistent signed direction across subjects or ROI pairs \cite{preti2017dynamic,yang2020reduced}. For boundary \(q\),
\begin{equation}
    \Delta\mathbf{FC}^{\mathrm{trn}}_{i,q}=\left|\mathbf{FC}^{\mathrm{scn}}_{i,q+1}-\mathbf{FC}^{\mathrm{scn}}_{i,q}\right|,
\label{eq:boundary_delta}
\end{equation}
where \(|\cdot|\) denotes elementwise absolute value. Here, \(q\) indexes the ordered narrative transition \(S_q\rightarrow S_{q+1}\), with each FC matrix summarizing its corresponding scene interval. The resulting tensor has shape \((S-1)\times R\times R\). Each ROI-level FC-delta row is encoded by a shared two-layer MLP, pooled across ROIs using mean, maximum, and standard deviation, and then aggregated across transitions with attention pooling to yield \(\mathbf{h}^{\mathrm{trn}}_i\).

\subsection{Global Context and Fusion}
\label{sec:global_contextual_features}

The global branch provides a whole-movie FC summary, a standard subject-level representation in connectome-based predictive modeling~\cite{shen2017using}. For subject \(i\), the same Ledoit--Wolf-based connectivity estimator is applied to the complete movie:
\begin{equation}
    \mathbf{FC}^{\mathrm{glb}}_i = \mathcal{Z}\!\left( \operatorname{Corr}_{\mathrm{LW}}(\mathbf{X}_i) \right), \quad
    \mathbf{x}^{\mathrm{glb}}_i = \operatorname{vec}_{\triangle} \!\left(\mathbf{FC}^{\mathrm{glb}}_i\right),
\label{eq:global_fc}
\end{equation}
where \(\operatorname{vec}_{\triangle}\) extracts off-diagonal upper-triangular entries, giving \(\mathbf{x}^{\mathrm{glb}}_i\in\mathbb{R}^{R(R-1)/2}\). A linear projection with dropout produces \(\mathbf{h}^{\mathrm{glb}}_i\). The final subject representation is the concatenation
\begin{equation}
    \mathbf{h}^{\mathrm{cat}}_i=
    \left[\mathbf{h}^{\mathrm{scn}}_i,
    \mathbf{h}^{\mathrm{trn}}_i,
    \mathbf{h}^{\mathrm{glb}}_i\right],\quad
    \mathbf{o}_i=f_{\mathrm{cls}}
    \left(\mathbf{h}^{\mathrm{cat}}_i\right),
\label{eq:concat_fusion}
\end{equation}
where \(f_{\mathrm{cls}}\) is an MLP classifier and
\(\mathbf{o}_i\) denotes the class logits. The model is trained
with class-weighted cross-entropy,
\begin{equation}
    \mathcal{L}=\operatorname{CE}_{w}\left(\mathbf{o}_i,y_i\right),
\label{eq:classification_loss}
\end{equation}
where \(y_i\) is the ground-truth label and class weights are
computed from the training data within each cross-validation fold.

\section{Experiments and Results}

\subsection{Dataset and Experiment Settings}

\begin{table}[t]
\caption{Demographics of the CMI-HBN cohort.}
\vspace{-2mm}
\centering
\renewcommand{\arraystretch}{0.9}
\setlength{\tabcolsep}{1.1pt}
\resizebox{0.85\linewidth}{!}{%
\begin{tabular}{@{}lccc@{}}
\toprule
\textbf{Group}
& \textbf{Subjects (N)}
& \textbf{Sex (M/F)}
& \textbf{Age (Mean$\pm$SD)} \\
\midrule

NoDx
& 71
& 40 / 31
& $8.99 \pm 1.82$ \\

ADHD-I
& 86
& 50 / 36
& $9.27 \pm 1.45$ \\

ADHD-C
& 103
& 54 / 49
& $8.98 \pm 1.76$ \\

\bottomrule
\end{tabular}%
}
\label{tab:cmihbn_demographics}
\vspace{-4mm}
\end{table}

~\noindent\textbf{Dataset.}
We used the publicly available Child Mind Institute Healthy Brain Network (CMI-HBN) \textit{Despicable Me} movie-fMRI dataset \cite{alexander2017open}. Participants aged 6--11 years whose structural MRI and blood-oxygen-level-dependent (BOLD) fMRI met the complete-pass quality control (QC) criteria were included, yielding 260 participants in the NoDx, ADHD-I, and ADHD-C groups (\autoref{tab:cmihbn_demographics}). Reproducible Brain Charts derivatives were generated with Configurable Pipeline for the Analysis of Connectomes (C-PAC) using standardized nuisance regression and temporal filtering \cite{shafiei2025reproducible,cpac2013}. QC required median framewise displacement (FD) of $\leq 0.2$ mm and template-registration normalized cross-correlation of $\geq 0.8$. The 10-min run contained 750 volumes (repetition time (TR)/echo time (TE) = 800/30 ms, flip angle = $31^{\circ}$, and 2.4-mm isotropic resolution); BOLD signals were parcellated into 100 Schaefer regions (SC-100) \cite{schaefer2018local}. Published adult-rater boundaries defined eight scenes \cite{cohen2022developmental}; each boundary was shifted by 4 s (five TRs) to approximate hemodynamic delay and mapped to the nearest TR \cite{lindquist2009modeling}. No statistically significant differences in median FD were detected across groups (Kruskal--Wallis, $p=0.092$) or between ADHD-I and ADHD-C (Mann--Whitney $U$, $p=0.078$).

\begin{table*}[!t]
\caption{
Performance comparison on CMI-HBN movie-fMRI ADHD classification. Best and second-best in \textbf{bold} and \underline{underline}.
}
\vspace{-4pt}
\centering
\renewcommand{\arraystretch}{1.1}
\setlength{\tabcolsep}{1.1pt}
\footnotesize

\resizebox{1\textwidth}{!}{%
\begin{tabular}{@{}lc cccc cccc ccccc@{}}
\toprule

\multirow[c]{2}{*}{\textbf{Model}}
&
\multirow[c]{2}{*}{\textbf{Venue}}
&
\multicolumn{4}{c}{\textbf{NoDx vs. ADHD}}
&
\multicolumn{4}{c}{\textbf{ADHD-I vs. ADHD-C}}
&
\multicolumn{5}{c}{\textbf{NoDx vs. ADHD-I vs. ADHD-C}} \\

\cmidrule(lr){3-6}
\cmidrule(lr){7-10}
\cmidrule(lr){11-15}

&
&
\makecell{\textbf{AUROC}\\$\uparrow$}
&
\makecell{\textbf{BACC}\\(\%) $\uparrow$}
&
\makecell{\textbf{SEN}\\(\%) $\uparrow$}
&
\makecell{\textbf{SPE}\\(\%) $\uparrow$}
&
\makecell{\textbf{AUROC}\\$\uparrow$}
&
\makecell{\textbf{BACC}\\(\%) $\uparrow$}
&
\makecell{\textbf{SEN}$_I$\\(\%) $\uparrow$}
&
\makecell{\textbf{SEN}$_C$\\(\%) $\uparrow$}
&
\makecell{\textbf{AUROC}\\$\uparrow$}
&
\makecell{\textbf{BACC}\\(\%) $\uparrow$}
&
\makecell{\textbf{SEN}$_N$\\(\%) $\uparrow$}
&
\makecell{\textbf{SEN}$_I$\\(\%) $\uparrow$}
&
\makecell{\textbf{SEN}$_C$\\(\%) $\uparrow$} \\

\midrule

SVM \cite{cortes1995support}
& ---
& \meanstd{0.53}{0.03}
& \meanstd{51.7}{1.9}
& \meanstd{63.0}{10.7}
& \meanstd{40.4}{18.0}
& \meanstd{0.61}{0.03}
& \meanstd{59.3}{0.5}
& \meanstd{48.1}{2.4}
& \meanstd{70.6}{2.4}
& \meanstd{0.67}{0.03}
& \meanstd{51.7}{2.5}
& \meanstd{41.3}{17.6}
& \bestmeanstd{63.2}{8.2}
& \meanstd{50.5}{17.2} \\

ED-HNN \cite{wang2022equivariant}
& ICLR'23
& \meanstd{0.62}{0.04}
& \meanstd{59.5}{4.0}
& \meanstd{68.3}{9.2}
& \meanstd{50.7}{11.0}
& \meanstd{0.67}{0.04}
& \meanstd{63.1}{2.2}
& \meanstd{52.3}{3.5}
& \meanstd{73.8}{1.0}
& \meanstd{0.68}{0.03}
& \meanstd{50.4}{6.8}
& \meanstd{44.7}{5.0}
& \underline{\meanstd{58.0}{3.5}}
& \meanstd{48.5}{14.6} \\

BioBGT \cite{ICLR2025_81f19c0e}
& ICLR'25
& \meanstd{0.53}{0.03}
& \meanstd{56.2}{2.1}
& \meanstd{46.7}{12.4}
& \bestmeanstd{65.7}{10.6}
& \meanstd{0.61}{0.03}
& \meanstd{59.5}{2.6}
& \meanstd{51.9}{5.7}
& \meanstd{67.0}{4.2}
& \meanstd{0.71}{0.01}
& \meanstd{55.1}{3.3}
& \meanstd{46.5}{2.4}
& \meanstd{55.4}{2.9}
& \bestmeanstd{63.4}{2.0} \\

DSAM \cite{thapaliya2025dsam}
& MedIA'25
& \meanstd{0.58}{0.02}
& \underline{\meanstd{60.4}{0.5}}
& \meanstd{61.2}{7.0}
& \meanstd{59.6}{7.8}
& \underline{\meanstd{0.70}{0.03}}
& \underline{\meanstd{65.8}{2.3}}
& \underline{\meanstd{62.8}{7.7}}
& \meanstd{68.9}{6.2}
& \meanstd{0.70}{0.01}
& \meanstd{53.3}{2.6}
& \meanstd{46.5}{2.4}
& \meanstd{55.8}{3.1}
& \meanstd{57.6}{3.7} \\

STNAGNN \cite{wang2025stnagnn}
& MIDL'25
& \meanstd{0.63}{0.02}
& \meanstd{58.7}{1.8}
& \meanstd{65.8}{9.7}
& \meanstd{51.6}{13.3}
& \meanstd{0.64}{0.08}
& \meanstd{59.3}{5.5}
& \meanstd{48.8}{11.6}
& \meanstd{69.9}{4.4}
& \underline{\meanstd{0.72}{0.01}}
& \meanstd{55.5}{1.3}
& \underline{\meanstd{51.2}{1.6}}
& \meanstd{55.8}{2.3}
& \meanstd{59.6}{5.0} \\

BrainHGT \cite{ma2026brainhgt}
& AAAI'26
& \underline{\meanstd{0.66}{0.02}}
& \meanstd{59.9}{4.7}
& \bestmeanstd{78.0}{6.2}
& \meanstd{41.8}{10.8}
& \meanstd{0.69}{0.01}
& \meanstd{64.3}{0.5}
& \meanstd{53.5}{3.5}
& \underline{\meanstd{75.1}{2.4}}
& \meanstd{0.71}{0.01}
& \underline{\meanstd{55.9}{2.8}}
& \meanstd{50.2}{9.6}
& \meanstd{56.2}{12.1}
& \meanstd{61.2}{1.9} \\ 

\midrule

\rowcolor{refpink!7}
\textbf{MovieSTAGE}
& ---
& \bestmeanstd{0.69}{0.04}
& \bestmeanstd{67.6}{3.9}
& \underline{\meanstd{69.8}{3.2}}
& \underline{\meanstd{65.3}{5.0}}
& \bestmeanstd{0.73}{0.04}
& \bestmeanstd{69.8}{2.2}
& \bestmeanstd{64.0}{5.1}
& \bestmeanstd{75.7}{1.0}
& \bestmeanstd{0.75}{0.03}
& \bestmeanstd{58.3}{3.1}
& \bestmeanstd{59.1}{9.7}
& \meanstd{53.5}{3.2}
& \underline{\meanstd{62.3}{8.6}} \\

\bottomrule
\end{tabular}%
}

\label{tab:benchmark_cmi_hbn}
\vspace{-4.5mm}
\end{table*}

~\noindent\textbf{Evaluation and Implementation.}
All methods shared the SC-100 parcellation, subject-level outer folds, and evaluation protocol. We performed 10 repetitions of stratified five-fold cross-validation. Standardization and class weights were fit within each outer-training fold; a class-stratified 20\% validation split supported a 12-configuration search, early stopping with patience 10 based on validation loss, and checkpoint selection by validation AUROC (one-vs-rest macro-AUROC for three classes), while outer test folds were reserved for final evaluation. For each repetition, test-fold predictions were concatenated into one complete out-of-fold (OOF) set, from which metrics were computed and summarized as mean \(\pm\) standard deviation across repetitions. AUROC and balanced accuracy (BACC) were primary, whereas sensitivity and specificity were descriptive. Paired differences were evaluated using class-stratified subject-cluster bootstrap 95\% confidence intervals and two-sided paired subject-cluster permutation tests (10,000 iterations each), with the mean repetition-wise metric difference as the statistic. Each subject's repeated predictions formed one bootstrap cluster and one joint permutation unit across repetitions. The same procedure was applied to the controlled three-class analyses, with Holm correction performed separately for baseline comparisons within each task and for the drop-one-branch, scene-only encoder, full-fusion encoder, and segmentation families across all AUROC and BACC contrasts (\(p_{\mathrm H}\)). MovieSTAGE was implemented in PyTorch and trained on an NVIDIA RTX A6000 GPU. Scene-wise hypergraphs used \(K=6\), and the Scene encoder comprised two hypergraph layers with a dropout rate of 0.1. The model was optimized for up to 80 epochs using AdamW (learning rate \(1.5\times10^{-4}\), weight decay \(2\times10^{-5}\), and batch size 16).

\subsection{Comparison with Baseline Methods}
\label{sec:comparison_graph_hypergraph}

We compared MovieSTAGE with a linear support vector machine (SVM) and five recent brain-network models. The SVM used the vectorized off-diagonal upper-triangular entries of whole-movie FC as input and served as a non-neural baseline \cite{cortes1995support}. STNAGNN modeled temporally ordered ROI graph snapshots using data-driven spatiotemporal attention \cite{wang2025stnagnn}. ED-HNN was adapted to subject-level ROI hypergraphs constructed from FC-profile similarity \cite{wang2022equivariant}. BioBGT and BrainHGT operated on whole-movie FC graphs \cite{ICLR2025_81f19c0e,ma2026brainhgt}, whereas DSAM directly processed the full ROI time series and learned connectivity through self-attention \cite{thapaliya2025dsam}. All baselines used the same SC-100 parcellation, subject-wise cross-validation splits, and evaluation metrics as MovieSTAGE.

As shown in \autoref{tab:benchmark_cmi_hbn}, MovieSTAGE achieved the highest mean AUROC and BACC point estimates across all three tasks. Paired comparisons of complete-OOF predictions further quantified the differences from the strongest competing baseline for each metric. For NoDx versus ADHD, MovieSTAGE exceeded BrainHGT by 0.03 in AUROC (95\% CI [0.006, 0.054], \(p_{\mathrm{H}}=0.028\)) and DSAM by 7.2 percentage points in BACC (95\% CI [2.6, 11.8], \(p_{\mathrm{H}}=0.008\)). For ADHD-I versus ADHD-C, MovieSTAGE exceeded DSAM by 0.03 in AUROC (95\% CI [0.005, 0.055], \(p_{\mathrm{H}}=0.036\)) and 4.0 percentage points in BACC (95\% CI [1.1, 6.9], \(p_{\mathrm{H}}=0.019\)). On the three-class task, MovieSTAGE exceeded STNAGNN by 0.03 in macro-AUROC (95\% CI [0.008, 0.052], \(p_{\mathrm{H}}=0.021\)) and BrainHGT by 2.4 percentage points in BACC (95\% CI [0.2, 4.6], \(p_{\mathrm{H}}=0.047\)). Although MovieSTAGE did not achieve the highest sensitivity for every individual class, it provided the strongest overall discrimination and class-balanced performance across the three tasks.

\begin{table}[t]
\caption{
Branch ablation on three-class ADHD classification.
}
\vspace{-4pt}
\centering
\renewcommand{\arraystretch}{1}
\setlength{\tabcolsep}{1pt}
\footnotesize

\resizebox{\columnwidth}{!}{%
\begin{tabular}{@{}ccc ccccc@{}}
\toprule

\multicolumn{3}{c}{\textbf{Branch configuration}}
&
\multicolumn{5}{c}{\textbf{Performance}} \\

\cmidrule(lr){1-3}
\cmidrule(lr){4-8}

\textbf{Scene}
& \textbf{Transition}
& \textbf{Global}
& \makecell{\textbf{AUROC}\\$\uparrow$}
& \makecell{\textbf{BACC}\\(\%) $\uparrow$}
& \makecell{\textbf{SEN}$_N$\\(\%) $\uparrow$}
& \makecell{\textbf{SEN}$_I$\\(\%) $\uparrow$}
& \makecell{\textbf{SEN}$_C$\\(\%) $\uparrow$} \\

\midrule

\cmark & \xmark & \xmark
& \meanstd{0.65}{0.01}
& \meanstd{46.7}{2.7}
& \meanstd{48.2}{7.9}
& \meanstd{49.8}{8.1}
& \meanstd{42.0}{8.8} \\

\xmark & \cmark & \xmark
& \meanstd{0.65}{0.01}
& \meanstd{47.2}{1.8}
& \meanstd{46.0}{6.2}
& \meanstd{46.1}{4.8}
& \meanstd{49.6}{5.2} \\

\xmark & \xmark & \cmark
& \meanstd{0.68}{0.03}
& \meanstd{49.4}{3.3}
& \meanstd{46.4}{12.4}
& \meanstd{55.4}{2.9}
& \meanstd{46.5}{7.6} \\

\midrule

\cmark & \cmark & \xmark
& \meanstd{0.68}{0.03}
& \meanstd{50.6}{2.7}
& \meanstd{48.7}{7.1}
& \meanstd{51.4}{3.2}
& \meanstd{51.8}{7.3} \\

\cmark & \xmark & \cmark
& \meanstd{0.71}{0.03}
& \meanstd{52.9}{4.9}
& \underline{\meanstd{52.7}{7.0}}
& \bestmeanstd{55.8}{2.9}
& \meanstd{50.1}{7.2} \\

\xmark & \cmark & \cmark
& \underline{\meanstd{0.71}{0.03}}
& \underline{\meanstd{53.3}{4.4}}
& \meanstd{51.6}{7.4}
& \underline{\meanstd{55.6}{2.5}}
& \underline{\meanstd{52.6}{7.1}} \\

\midrule

\rowcolor{refpink!7}
\cmark & \cmark & \cmark
& \bestmeanstd{0.75}{0.03}
& \bestmeanstd{58.3}{3.1}
& \bestmeanstd{59.1}{9.7}
& \meanstd{53.5}{3.2}
& \bestmeanstd{62.3}{8.6} \\

\bottomrule
\end{tabular}%
}

\label{tab:branch_ablation_multiclass}
\vspace{-4mm}
\end{table}

\subsection{Branch Ablation}



As shown in \autoref{tab:branch_ablation_multiclass}, Global was the strongest single branch, whereas the full model achieved a macro-AUROC of \(0.75 \pm 0.03\) and a BACC of \(58.3 \pm 3.1\%\). In Holm-corrected drop-one-branch comparisons, the full model outperformed all three two-branch variants for both primary metrics (all \(p_{\mathrm H}\leq0.041\)). Relative to Transition+Global, the strongest two-branch configuration, the gains were \(0.04\) in macro-AUROC (95\% CI [0.009, 0.071], \(p_{\mathrm H}=0.031\)) and \(5.0\) percentage points in BACC (95\% CI [1.1, 8.9], \(p_{\mathrm H}=0.026\)). Overall, the ablation supports conditional and partially complementary contributions from within-scene organization, adjacent-scene reconfiguration, and global FC context, rather than uniformly strong independent contributions from each branch.

\begin{figure*}[!t]
    \centering
    \includegraphics[width=1\textwidth]{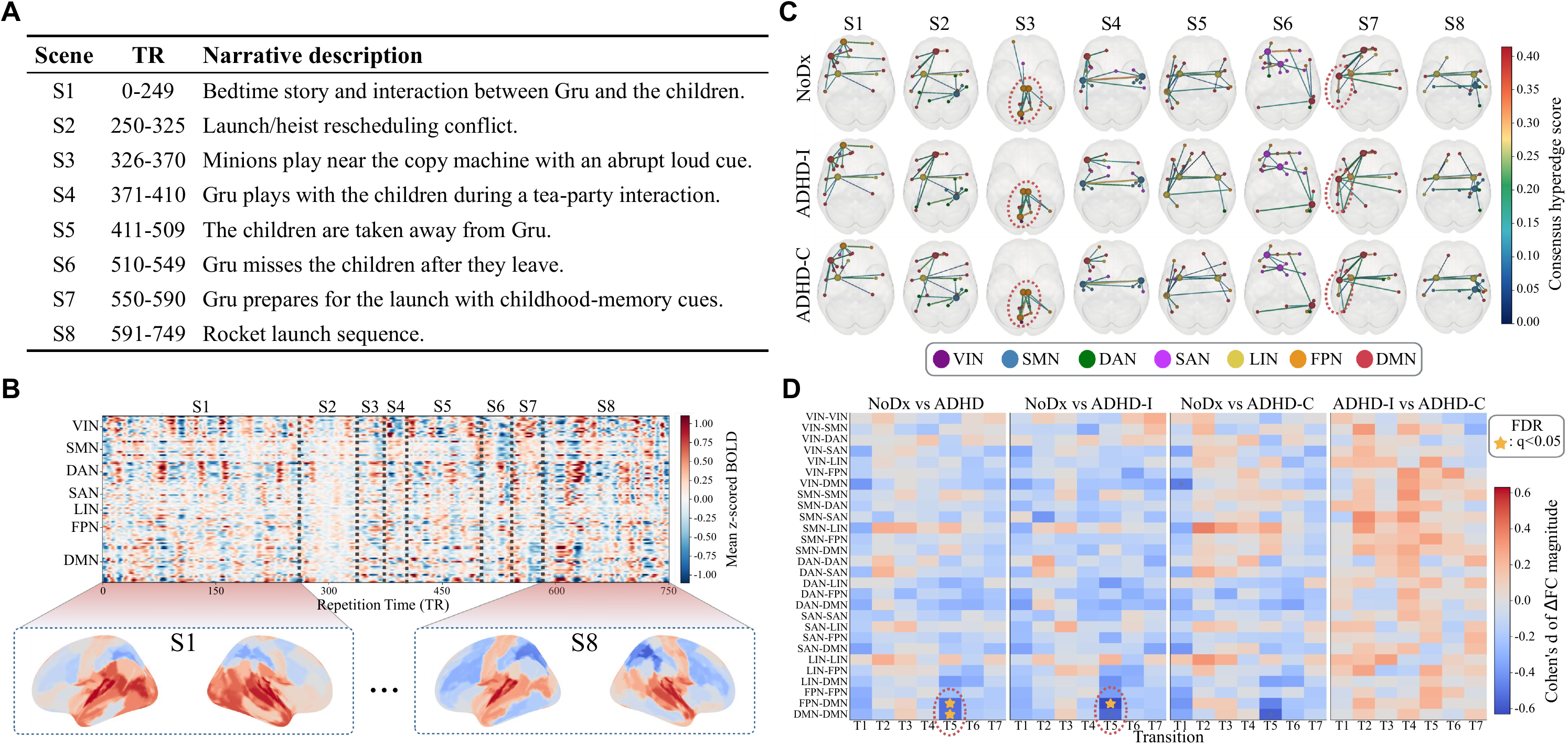}
    \caption{Post-hoc interpretability analyses based on the Yeo 7-network parcellation~\cite{thomas2011organization}: Visual (\textcolor{yeoVis}{\textbf{VIN}}), Somatomotor (\textcolor{yeoSomMot}{\textbf{SMN}}), Dorsal Attention (\textcolor{yeoDorsAttn}{\textbf{DAN}}), Salience/Ventral Attention (\textcolor{yeoSalVentAttn}{\textbf{SAN}}), Limbic (\textcolor{yeoLimbic}{\textbf{LIN}}),  Frontoparietal (\textcolor{yeoCont}{\textbf{FPN}}), and Default Mode (\textcolor{yeoDefault}{\textbf{DMN}}). (A) TR intervals and narrative descriptions of the eight predefined movie scenes (S1--S8). (B) Group-averaged, within-ROI z-scored SC-100 BOLD response profiles segmented into S1--S8 movie scenes. (C) Group-level scene-wise consensus hypergraphs for NoDx, ADHD-I, and ADHD-C; tubes denote model-derived anchor-to-member hyperedge projections, not direct pairwise physiological connectivity. (D) Network-pair Cohen's \(d\) heatmaps of adjacent-scene \(\Delta\)FC reconfiguration for T1--T7. Markers indicate permutation-based BH-FDR results (star, \(q<0.05\)). Cohen's \(d\) signs indicate group differences in reconfiguration magnitude, not signed FC changes.}
    \label{fig:Interpretability_Analysis}
    \vspace{-4mm}
\end{figure*}

\begin{table}[t]
\caption{
Three-class ADHD classification across scene encoders.
}
\label{tab:scene_aggregation}
\vspace{-4pt}
\centering
\setlength{\tabcolsep}{1.5pt}
\renewcommand{\arraystretch}{1.}
\footnotesize
\resizebox{\columnwidth}{!}{%
\begin{tabular}{lcccc}
\toprule
\multirow{2}{*}{\textbf{Scene encoder}}
& \multicolumn{2}{c}{\textbf{Scene only}}
& \multicolumn{2}{c}{\textbf{Full fusion}} \\
\cmidrule(lr){2-3}
\cmidrule(lr){4-5}
& \textbf{AUROC} $\uparrow$
& \textbf{BACC (\%)} $\uparrow$
& \textbf{AUROC} $\uparrow$
& \textbf{BACC (\%)} $\uparrow$ \\
\midrule
MLP
& \meanstd{0.60}{0.01} & \meanstd{43.1}{2.9} & \meanstd{0.69}{0.02} & \meanstd{52.1}{2.8} \\
GAT
& \underline{\meanstd{0.63}{0.01}} & \meanstd{45.0}{3.2} & \underline{\meanstd{0.71}{0.02}} & \underline{\meanstd{53.2}{4.4}} \\
BNT
& \meanstd{0.61}{0.01} & \underline{\meanstd{45.2}{1.7}} & \meanstd{0.70}{0.02} & \meanstd{52.1}{3.0} \\

\rowcolor{refpink!7}
\textbf {HGNN (MovieSTAGE)}
& \bestmeanstd{0.65}{0.01}
& \bestmeanstd{46.7}{2.7}
& \bestmeanstd{0.75}{0.03}
& \bestmeanstd{58.3}{3.1} \\
\bottomrule
\end{tabular}%
}
\vspace{-4mm}
\end{table}

\subsection{Controlled Comparison of Scene Encoders}
\label{sec:scene_encoder_comparison}

On the three-class task, we assessed the contribution of explicit hypergraph-structured aggregation to within-scene FC-profile modeling by replacing MovieSTAGE's hypergraph neural network (HGNN) scene encoder with an ROI-wise multilayer perceptron (MLP), a graph attention network (GAT) \cite{2018graph}, and a Brain Network Transformer (BNT) \cite{kan2022brain}. All variants used identical human-annotated segmentation and scene-wise FC-profile node features and shared the same ROI-level readout, scene-embedding dimensionality, cross-scene attention pooling, subject-level cross-validation splits, training and evaluation protocol, internal-validation procedure, and hyperparameter-search budget; only the within-scene encoder was replaced. In the full-fusion setting, the Transition and Global branches and the fusion classifier were unchanged and jointly optimized with each scene encoder. Each encoder was evaluated in both the scene-only and full-fusion settings to assess its performance as a standalone scene representation and as a component of the complete multibranch model.

As shown in \autoref{tab:scene_aggregation}, the HGNN yielded the highest mean macro-AUROC and BACC in both settings. In Holm-corrected comparisons, it outperformed all three alternative encoders for both metrics in the scene-only and full-fusion settings (scene only: all \(p_{\mathrm H}\leq0.047\); full fusion: all \(p_{\mathrm H}\leq0.027\)). In the scene-only setting, the gains over the strongest alternative for each metric were \(0.02\) in macro-AUROC relative to GAT (95\% CI [0.004, 0.036], \(p_{\mathrm H}=0.041\)) and \(1.5\) percentage points in BACC relative to BNT (95\% CI [0.1, 2.9], \(p_{\mathrm H}=0.047\)). Under full fusion, the HGNN exceeded GAT by \(0.04\) in macro-AUROC (95\% CI [0.010, 0.070], \(p_{\mathrm H}=0.027\)) and \(5.1\) percentage points in BACC (95\% CI [1.2, 9.0], \(p_{\mathrm H}=0.021\)). Overall, under matched scene inputs, readout operations, embedding dimensionality, tuning budget, and downstream components, these results were consistent with a benefit of the explicit hypergraph inductive bias for modeling within-scene FC-profile organization among the evaluated encoders.


\subsection{Controlled Comparison of Fixed-Count Scene Partitions}

\begin{table}[t]
\caption{
Controlled comparison of eight-segment scene partitions.
}
\vspace{-4pt}
\centering
\renewcommand{\arraystretch}{1}
\setlength{\tabcolsep}{1.5pt}
\footnotesize

\resizebox{\columnwidth}{!}{%
\begin{tabular}{@{}l ccccc@{}}
\toprule

\textbf{Method}
& \makecell{\textbf{AUROC}\\$\uparrow$}
& \makecell{\textbf{BACC}\\(\%) $\uparrow$}
& \makecell{\textbf{SEN}$_N$\\(\%) $\uparrow$}
& \makecell{\textbf{SEN}$_I$\\(\%) $\uparrow$}
& \makecell{\textbf{SEN}$_C$\\(\%) $\uparrow$} \\

\midrule

Random
& \meanstd{0.70}{0.03}
& \meanstd{53.2}{2.3}
& \meanstd{50.7}{13.5}
& \meanstd{53.4}{2.6}
& \underline{\meanstd{55.4}{9.6}} \\

GSBS
& \underline{\meanstd{0.71}{0.06}}
& \underline{\meanstd{55.8}{1.8}}
& \underline{\meanstd{56.2}{13.2}}
& \bestmeanstd{61.6}{2.9}
& \meanstd{49.6}{6.8} \\

\rowcolor{refpink!7}
Human
& \bestmeanstd{0.75}{0.03}
& \bestmeanstd{58.3}{3.1}
& \bestmeanstd{59.1}{9.7}
& \underline{\meanstd{53.5}{3.2}}
& \bestmeanstd{62.3}{8.6} \\

\bottomrule
\end{tabular}%
}

\label{tab:scene_segmentation_comparison}
\vspace{-5mm}
\end{table}

On the three-class task, we compared three eight-segment partitions under an otherwise matched downstream protocol to assess the effect of boundary placement: human-annotated narrative scenes, random partitions matched in segment count and duration distribution, and GSBS-derived neural-state segments constrained to eight states
\cite{geerligs2021detecting}. Matching the segment count fixed the number of scene and transition representations; thus, the GSBS condition serves as a fixed-count boundary-placement control rather than a benchmark of unconstrained GSBS. One independent random partition was generated per cross-validation repetition, whereas GSBS boundaries were re-estimated within each outer fold using only the corresponding outer-training subjects. The model architecture, hyperparameter search space and budget, subject splits, training procedure, and complete-OOF evaluation protocol were otherwise identical. As shown in \autoref{tab:scene_segmentation_comparison}, human-annotated segmentation yielded the highest mean macro-AUROC (\(0.75 \pm 0.03\)) and BACC (\(58.3 \pm 3.1\%\)) and outperformed both controls for both primary metrics after Holm correction (all \(p_{\mathrm H}\leq0.047\)). Relative to Random, it improved macro-AUROC by \(0.05\) (95\% CI [0.014, 0.086], \(p_{\mathrm H}=0.015\)) and BACC by \(5.1\) percentage points (95\% CI [1.3, 8.9], \(p_{\mathrm H}=0.018\)); relative to GSBS, the corresponding gains were \(0.04\) (95\% CI [0.006, 0.074], \(p_{\mathrm H}=0.039\)) and \(2.5\) percentage points (95\% CI [0.2, 4.8], \(p_{\mathrm H}=0.047\)). The Human--Random comparison was therefore consistent with a benefit of alignment to the annotated narrative-event structure beyond segment count and duration matching. Although GSBS yielded higher ADHD-I sensitivity, human-annotated segmentation provided higher overall discrimination and class-balanced performance at the level of mean point estimates.

\subsection{Post-hoc Interpretability Analyses}
\label{sec:interpretability_analysis}

\autoref{fig:Interpretability_Analysis} summarizes the post-hoc interpretability analyses. \autoref{fig:Interpretability_Analysis}A lists the TR intervals and narrative content of the eight predefined movie scenes, providing the temporal reference for the scene-wise and transition-level analysis. Scenes are denoted S1--S8, and adjacent boundaries are denoted T1--T7, where \(T_q\) is the transition from \(S_q\) to \(S_{q+1}\). \autoref{fig:Interpretability_Analysis}B shows group-averaged, within-ROI z-scored SC-100 BOLD response profiles across the eight movie scenes. The profiles varied across scenes, indicating that the movie elicited temporally heterogeneous cortical response contexts rather than a homogeneous whole-run pattern. For example, S3 and the S5--S6 transition showed visible changes in network-level BOLD modulation, providing descriptive context for the scene-wise and transition-level analyses.

Scene-wise consensus hypergraphs (\autoref{fig:Interpretability_Analysis}C) were obtained by averaging learned hyperedge weights across subjects within each group, with high-weight anchors defined by the highest averaged hyperedge weights. Across scenes, selected anchors were distributed across FPN, DMN, LIN, SMN, and SAN, indicating that the learned scene representation emphasized multiple large-scale systems implicated in ADHD. S3, a salient audiovisual copy-machine scene, exhibited an FPN-centered topology, whereas S7, involving launch preparation and childhood-memory cues, exhibited a DMN/LIN-centered topology. These examples characterize model-derived topology descriptively; formal diagnostic-group inference is reported only for the transition-level analysis, and the hyperedge projections do not represent direct physiological connectivity \cite{tegelbeckers2015altered,salmi2020adhd,fair2010atypical}.

Transition-level analysis (\autoref{fig:Interpretability_Analysis}D) showed the strongest effects for T5, the transition from the children leaving Gru to Gru missing them. For this adjacent-scene contrast, the pooled ADHD group showed greater reconfiguration magnitude than NoDx for FPN--DMN and DMN--DMN network pairs after permutation-based Benjamini–Hochberg false-discovery-rate (BH-FDR) correction; ADHD-I also showed greater FPN--DMN reconfiguration than NoDx. Because \(\Delta\)FC is an absolute adjacent-scene difference, these effects indicate group differences in reconfiguration magnitude rather than signed connectivity increases or decreases. The concentration of DMN-centered and FPN--DMN effects aligns with ADHD literature on reduced task-positive/task-negative segregation, increased DMN integration with task-relevant networks, and altered dynamic connectivity states \cite{mills2018adhd,duffy2021increased,shappell2021children}.

\section{Discussion}
\label{sec:discussion}

We examined whether narrative-aligned representations retain predictive information beyond whole-movie FC during naturalistic viewing. Across the case--control, subtype, and three-class CMI-HBN tasks, the full MovieSTAGE model yielded the highest mean AUROC and BACC point estimates among the evaluated methods. Although whole-movie FC was the strongest individual representation, combining it with within-scene organization and adjacent-scene reconfiguration produced the highest overall performance. The HGNN also yielded the highest point estimates among the evaluated scene encoders, while human-annotated scenes yielded higher point estimates than duration-matched random and fixed-count GSBS partitions. Together, these controlled analyses suggest that whole-movie FC provides broad subject-level context, whereas the Scene and Transition branches retain event-specific information not fully captured by global summaries or temporal partitioning alone. This interpretation is consistent with prior connectome-based prediction and with evidence that continuous narratives comprise relatively stable cortical event patterns separated by transition-related shifts~\cite{shen2017using,finn2021movie,baldassano2017discovering,cohen2022developmental,lee2024relationship}.

The post-hoc analyses placed these findings in a network-level context. Scene-wise hypergraphs emphasized FPN, DMN, LIN, SMN, and SAN systems, whereas the strongest corrected transition-level differences involved FPN--DMN and DMN--DMN reconfiguration at the S5--S6 transition. These patterns align with prior ADHD findings involving default-mode organization, task-positive/task-negative segregation, salience processing, and dynamic network states~\cite{tegelbeckers2015altered,fair2010atypical,mills2018adhd,duffy2021increased,shappell2021children}. They should nevertheless be interpreted as model-derived associations: hyperedge projections do not represent direct physiological connectivity, and absolute \(\Delta\)FC reflects reconfiguration magnitude rather than signed connectivity change. The main limitations are evaluation within a single pediatric cohort and unequal temporal support across predefined scenes. External validation on independent movie-fMRI cohorts and robustness analyses across longer stimuli and alternative event definitions are needed. Within these bounds, the results support event-aligned multiscale modeling as a useful framework for ADHD prediction and for generating network-level hypotheses about scene- and transition-related functional organization.

\section{Conclusion}
\label{sec:conclusion}

We introduced MovieSTAGE, an event-aligned scene--transition--global framework combining scene-specific hypergraph modeling, adjacent-scene FC-profile reconfiguration, and whole-movie FC. Across three CMI-HBN tasks, MovieSTAGE achieved the highest mean AUROC and BACC among the evaluated methods. Three-class branch ablations indicated conditional and partially complementary contributions. Post-hoc scene-wise hypergraph and transition analyses generated network-level hypotheses about scene-dependent functional organization and adjacent-scene reconfiguration. These results support event-aligned multiscale representation learning for naturalistic fMRI, while independent external validation remains necessary.

\bibliographystyle{IEEEtran}
\bibliography{ref}

@article{finn2021movie,
  title={Movie-watching outperforms rest for functional connectivity-based prediction of behavior},
  author={Finn, Emily S and Bandettini, Peter A},
  journal={NeuroImage},
  volume={235},
  pages={117963},
  year={2021},
  publisher={Elsevier}
}

@article{alexander2017open,
  title={An open resource for transdiagnostic research in pediatric mental health and learning disorders},
  author={Alexander, Lindsay M and Escalera, Jasmine and Ai, Lei and Andreotti, Charissa and Febre, Karina and Mangone, Alexander and Vega-Potler, Natan and Langer, Nicolas and Alexander, Alexis and Kovacs, Meagan and others},
  journal={Scientific Data},
  volume={4},
  number={1},
  pages={170181},
  year={2017},
  publisher={Nature Publishing Group}
}

@article{baldassano2017discovering,
  title={Discovering event structure in continuous narrative perception and memory},
  author={Baldassano, Christopher and Chen, Janice and Zadbood, Asieh and Pillow, Jonathan W and Hasson, Uri and Norman, Kenneth A},
  journal={Neuron},
  volume={95},
  number={3},
  pages={709--721},
  year={2017},
  publisher={Elsevier}
}

@article{schaefer2018local,
  title={Local-global parcellation of the human cerebral cortex from intrinsic functional connectivity {MRI}},
  author={Schaefer, Alexander and Kong, Ru and Gordon, Evan M and Laumann, Timothy O and Zuo, Xi-Nian and Holmes, Avram J and Eickhoff, Simon B and Yeo, BT Thomas},
  journal={Cerebral Cortex},
  volume={28},
  number={9},
  pages={3095--3114},
  year={2018},
  publisher={Oxford University Press}
}

@article{cohen2022developmental,
  title={Developmental changes in story-evoked responses in the neocortex and hippocampus},
  author={Cohen, Samantha S and Tottenham, Nim and Baldassano, Christopher},
  journal={eLife},
  volume={11},
  pages={e69430},
  year={2022},
  publisher={eLife Sciences Publications, Ltd}
}

@article{shafiei2025reproducible,
  title={Reproducible Brain Charts: An open data resource for mapping brain development and its associations with mental health},
  author={Shafiei, Golia and Esper, Nathalia B and Hoffmann, Mauricio S and Ai, Lei and Chen, Andrew A and Cluce, Jon and Covitz, Sydney and Giavasis, Steven and Lane, Connor and Mehta, Kahini and others},
  journal={Neuron},
  volume={113},
  number={22},
  pages={3758--3779},
  year={2025},
  publisher={Elsevier}
}

@article{cortes1995support,
  title={Support-vector networks},
  author={Cortes, Corinna and Vapnik, Vladimir},
  journal={Machine Learning},
  volume={20},
  number={3},
  pages={273--297},
  year={1995},
  publisher={Springer}
}

@inproceedings{feng2019hypergraph,
  title={Hypergraph neural networks},
  author={Feng, Yifan and You, Haoxuan and Zhang, Zizhao and Ji, Rongrong and Gao, Yue},
  booktitle={Proceedings of the AAAI conference on artificial intelligence},
  volume={33},
  number={01},
  pages={3558--3565},
  year={2019}
}

@article{thapaliya2025dsam,
  title={{DSAM}: A deep learning framework for analyzing temporal and spatial dynamics in brain networks},
  author={Thapaliya, Bishal and Miller, Robyn and Chen, Jiayu and Wang, Yu Ping and Akbas, Esra and Sapkota, Ram and Ray, Bhaskar and Suresh, Pranav and Ghimire, Santosh and Calhoun, Vince D and others},
  journal={Medical Image Analysis},
  volume={101},
  pages={103462},
  year={2025},
  publisher={Elsevier}
}

@inproceedings{ma2026brainhgt,
  title={{BrainHGT}: A Hierarchical Graph Transformer for Interpretable Brain Network Analysis},
  author={Ma, Jiajun and Zhang, Yongchao and Zhang, Chao and Lv, Zhao and Pei, Shengbing},
  booktitle={Proceedings of the AAAI Conference on Artificial Intelligence},
  volume={40},
  number={21},
  pages={17617--17625},
  year={2026}
}

@article{shen2017using,
  title={Using connectome-based predictive modeling to predict individual behavior from brain connectivity},
  author={Shen, Xilin and Finn, Emily S and Scheinost, Dustin and Rosenberg, Monica D and Chun, Marvin M and Papademetris, Xenophon and Constable, R Todd},
  journal={Nature Protocols},
  volume={12},
  number={3},
  pages={506--518},
  year={2017},
  publisher={Nature Publishing Group UK London}
}

@article{lee2024relationship,
  title={The relationship between event boundary strength and pattern shifts across the cortical hierarchy during naturalistic movie-viewing},
  author={Lee, Yoonjung and Chen, Janice},
  journal={Journal of Cognitive Neuroscience},
  volume={36},
  number={11},
  pages={2317--2342},
  year={2024},
  publisher={MIT Press 255 Main Street, 9th Floor, Cambridge, Massachusetts 02142, USA~…}
}

@article{xiao2019multi,
  title={Multi-hypergraph learning-based brain functional connectivity analysis in {fMRI} data},
  author={Xiao, Li and Wang, Junqi and Kassani, Peyman H and Zhang, Yipu and Bai, Yuntong and Stephen, Julia M and Wilson, Tony W and Calhoun, Vince D and Wang, Yu-Ping},
  journal={IEEE Transactions on Medical Imaging},
  volume={39},
  number={5},
  pages={1746--1758},
  year={2019},
  publisher={IEEE}
}

@article{tegelbeckers2015altered,
  title={Altered salience processing in attention deficit hyperactivity disorder},
  author={Tegelbeckers, Jana and Bunzeck, Nico and Duzel, Emrah and Bonath, Bj{\"o}rn and Flechtner, Hans-Henning and Krauel, Kerstin},
  journal={Human Brain Mapping},
  volume={36},
  number={6},
  pages={2049--2060},
  year={2015},
  publisher={Wiley Online Library}
}

@article{salmi2020adhd,
  title={{ADHD} desynchronizes brain activity during watching a distracted multi-talker conversation},
  author={Salmi, Juha and Metwaly, Mostafa and Tohka, Jussi and Alho, Kimmo and Lepp{\"a}m{\"a}ki, Sami and Tani, Pekka and Koski, Anniina and Vanderwal, Tamara and Laine, Matti},
  journal={NeuroImage},
  volume={216},
  pages={116352},
  year={2020},
  publisher={Elsevier}
}

@article{mills2018adhd,
  title={{ADHD} and attentional control: Impaired segregation of task positive and task negative brain networks},
  author={Mills, Brian D and Miranda-Dominguez, Oscar and Mills, Kathryn L and Earl, Eric and Cordova, Michaela and Painter, Julia and Karalunas, Sarah L and Nigg, Joel T and Fair, Damien A},
  journal={Network Neuroscience},
  volume={2},
  number={02},
  pages={200--217},
  year={2018},
  publisher={MIT Press One Rogers Street, Cambridge, MA 02142-1209, USA journals-info~…}
}

@article{duffy2021increased,
  title={Increased integration between default mode and task-relevant networks in children with {{ADHD}} is associated with impaired response control},
  author={Duffy, Kelly A and Rosch, Keri S and Nebel, Mary Beth and Seymour, Karen E and Lindquist, Martin A and Pekar, James J and Mostofsky, Stewart H and Cohen, Jessica R},
  journal={Developmental Cognitive Neuroscience},
  volume={50},
  pages={100980},
  year={2021},
  publisher={Elsevier}
}

@article{shappell2021children,
  title={Children with attention-deficit/hyperactivity disorder spend more time in hyperconnected network states and less time in segregated network states as revealed by dynamic connectivity analysis},
  author={Shappell, Heather M and Duffy, Kelly A and Rosch, Keri S and Pekar, James J and Mostofsky, Stewart H and Lindquist, Martin A and Cohen, Jessica R},
  journal={NeuroImage},
  volume={229},
  pages={117753},
  year={2021},
  publisher={Elsevier}
}

@article{fair2010atypical,
  title={Atypical default network connectivity in youth with attention-deficit/hyperactivity disorder},
  author={Fair, Damien A and Posner, Jonathan and Nagel, Bonnie J and Bathula, Deepti and Dias, Taciana G Costa and Mills, Kathryn L and Blythe, Michael S and Giwa, Aishat and Schmitt, Colleen F and Nigg, Joel T},
  journal={Biological Psychiatry},
  volume={68},
  number={12},
  pages={1084--1091},
  year={2010},
  publisher={Elsevier}
}

@article{savva2019assessment,
  title={Assessment of dynamic functional connectivity in resting-state {fMRI} using the sliding window technique},
  author={Savva, Antonis D and Mitsis, Georgios D and Matsopoulos, George K},
  journal={Brain and Behavior},
  volume={9},
  number={4},
  pages={e01255},
  year={2019},
  publisher={Wiley Online Library}
}

@article{lindquist2009modeling,
  title={Modeling the hemodynamic response function in {fMRI}: efficiency, bias and mis-modeling},
  author={Lindquist, Martin A and Loh, Ji Meng and Atlas, Lauren Y and Wager, Tor D},
  journal={NeuroImage},
  volume={45},
  number={1},
  pages={S187--S198},
  year={2009},
  publisher={Elsevier}
}

@article{geerligs2021detecting,
  title={Detecting neural state transitions underlying event segmentation},
  author={Geerligs, Linda and van Gerven, Marcel and G{\"u}{\c{c}}l{\"u}, Umut},
  journal={NeuroImage},
  volume={236},
  pages={118085},
  year={2021},
  publisher={Elsevier}
}

@article{preti2017dynamic,
  title={The dynamic functional connectome: State-of-the-art and perspectives},
  author={Preti, Maria Giulia and Bolton, Thomas AW and Van De Ville, Dimitri},
  journal={NeuroImage},
  volume={160},
  pages={41--54},
  year={2017},
  publisher={Elsevier}
}

@article{ledoit2004well,
  title={A well-conditioned estimator for large-dimensional covariance matrices},
  author={Ledoit, Olivier and Wolf, Michael},
  journal={Journal of Multivariate Analysis},
  volume={88},
  number={2},
  pages={365--411},
  year={2004},
  publisher={Elsevier}
}

@article{kan2022brain,
  title={Brain network transformer},
  author={Kan, Xuan and Dai, Wei and Cui, Hejie and Zhang, Zilong and Guo, Ying and Yang, Carl},
  journal={Advances in Neural Information Processing Systems},
  volume={35},
  pages={25586--25599},
  year={2022}
}

@inproceedings{cao2021temporal,
  title={Temporal graph representation learning for autism spectrum disorder brain networks},
  author={Cao, Peng and Wen, Guangqi and Li, Lanting and Liu, Xiaoli and Yang, Jinzhu and Zaiane, Osmar},
  booktitle={2021 IEEE International Conference on Bioinformatics and Biomedicine (BIBM)},
  pages={1270--1275},
  year={2021},
  organization={IEEE}
}

@inproceedings{yao2014functional,
  title={Functional network disruption in attention deficit hyperactivity disorder},
  author={Yao, Zhijun and Hu, Bin and Xie, Yuanwei and Wang, Wei and Liu, Ruiyue and Liang, Chuanjiang and Su, Yun},
  booktitle={2014 IEEE International Conference on Bioinformatics and Biomedicine (BIBM)},
  pages={84--90},
  year={2014},
  organization={IEEE}
}

@inproceedings{yang2020reduced,
  title={Reduced dynamics in multivariate regression-based dynamic connectivity of depressive disorder},
  author={Yang, Zhengwu and Guo, Hanning and Ji, Shanling and Li, Shan and Fu, Yu and Guo, Man and Yao, Zhijun},
  booktitle={2020 IEEE International Conference on Bioinformatics and Biomedicine (BIBM)},
  pages={1197--1201},
  year={2020},
  organization={IEEE}
}

@inproceedings{wang2022equivariant,
  title={Equivariant Hypergraph Diffusion Neural Operators},
  author={Wang, Peihao and Yang, Shenghao and Liu, Yunyu and Wang, Zhangyang and Li, Pan},
  booktitle={International Conference on Learning Representations (ICLR)},
  year={2023}
}

@inproceedings{ICLR2025_81f19c0e,
 author = {Peng, Ciyuan and Huang, Yuelong and Dong, Qichao and Yu, Shuo and Xia, Feng and Zhang, Chengqi and Jin, Yaochu},
 booktitle = {International Conference on Learning Representations},
 editor = {Y. Yue and A. Garg and N. Peng and F. Sha and R. Yu},
 pages = {52283--52309},
 title = {Biologically Plausible Brain Graph Transformer},
 url = {https://proceedings.iclr.cc/paper_files/paper/2025/file/81f19c0e9f3e06c831630ab6662fd8ea-Paper-Conference.pdf},
 volume = {2025},
 year = {2025}
}

@inproceedings{2018graph,
title={Graph Attention Networks},
author={Petar Veličković and Guillem Cucurull and Arantxa Casanova and Adriana Romero and Pietro Liò and Yoshua Bengio},
booktitle={International Conference on Learning Representations},
year={2018},
url={https://openreview.net/forum?id=rJXMpikCZ},
}

@ARTICLE{cpac2013,
    AUTHOR={Craddock, Cameron  and  Sikka, Sharad  and  Cheung, Brian  and  Khanuja, Ranjeet  and  Ghosh, Satrajit S
        and Yan, Chaogan  and  Li, Qingyang  and  Lurie, Daniel  and  Vogelstein, Joshua  and  Burns, Randal  and
        Colcombe, Stanley  and  Mennes, Maarten  and  Kelly, Clare  and  Di Martino, Adriana  and  Castellanos,
        Francisco Xavier  and  Milham, Michael},
    TITLE={Towards Automated Analysis of Connectomes: The Configurable Pipeline for the Analysis of Connectomes ({C-PAC})},
    JOURNAL={Frontiers in Neuroinformatics},
    YEAR={2013},
    NUMBER={42},
    DOI={10.3389/conf.fninf.2013.09.00042},
    ISSN={1662-5196}
}

@inproceedings{
wang2025stnagnn,
title={{STNAGNN}: Data-driven Spatio-temporal Brain Connectivity beyond {FC}},
author={Jiyao Wang and Nicha C Dvornek and Peiyu Duan and Lawrence H. Staib and Pamela Ventola and James S Duncan},
booktitle={Medical Imaging with Deep Learning},
year={2025},
url={https://openreview.net/forum?id=nKLCB8d3Ko}
}

@article{thomas2011organization,
  title={The organization of the human cerebral cortex estimated by intrinsic functional connectivity},
  author={Thomas Yeo, BT and Krienen, Fenna M and Sepulcre, Jorge and Sabuncu, Mert R and Lashkari, Danial and Hollinshead, Marisa and Roffman, Joshua L and Smoller, Jordan W and Z{\"o}llei, Lilla and Polimeni, Jonathan R and others},
  journal={Journal of neurophysiology},
  volume={106},
  number={3},
  pages={1125--1165},
  year={2011},
  publisher={American Physiological Society Bethesda, MD}
}

\end{document}